\def\year{2016}\relax
\documentclass[journal,transmag]{IEEEtran}

\usepackage{times}
\usepackage{helvet}
\usepackage{courier}

\usepackage{graphicx}
\usepackage{amsmath,amssymb} 
\usepackage{color}
\usepackage{times}
\usepackage{epsfig}
\usepackage{float}
\usepackage{url}
\usepackage{subfig}
\usepackage{multirow}
\usepackage{algorithm}
\usepackage{algpseudocode}

\usepackage{array}
\newcolumntype{x}[1]{>{\centering\arraybackslash\hspace{0pt}}m{#1}}
\usepackage{epstopdf}
\newcommand{\KG}{\textcolor{black}}
\newcommand{\SJ}{\textcolor{black}}
\newcommand{\SDJ}{\textcolor{black}}

\begin{document}
%
\title{Click Carving: Segmenting Objects in Video with Point Clicks} 

\author{
	\IEEEauthorblockN{Suyog~Dutt~Jain and
		Kristen~Grauman}
	\IEEEauthorblockA{Department of Computer Science \\ The University of Texas at Austin}
}

\maketitle

\begin{abstract}
We present a novel form of interactive video object segmentation where a few clicks by the user helps the system produce a full spatio-temporal segmentation \SDJ{of the object of interest}.  Whereas conventional interactive pipelines take the user's initialization as a starting point, we show the value in the system taking the lead even in initialization. \SDJ{ In particular, for a given video frame, the system precomputes a ranked list of thousands of possible segmentation hypotheses (also referred to as object region proposals) using image and motion cues.} Then, the user looks at the top ranked proposals, and clicks on the object boundary to carve away erroneous ones.  This process iterates (typically 2-3 times), and each time the system revises the top ranked proposal set, until the user is satisfied with a resulting segmentation mask.  Finally, the mask is propagated across the video to produce a spatio-temporal object tube.  On three challenging datasets, we provide extensive comparisons with both existing work and simpler alternative methods.  In all, the proposed Click Carving approach strikes an excellent balance of accuracy and human effort.  It outperforms all similarly fast methods, and is competitive or better than those requiring 2 to 12 times the effort.

\end{abstract}
\section{Introduction}\label{sec:introduction}



Video object segmentation entails computing a pixel-level mask for an object(s) across the frames of video, regardless of that object's category.  Analogous to image segmentation, which produces a 2D map delineating the object's spatial region (``blob"), video segmentation produces a 3D map delineating the object's spatio-temporal extent (``tube").  The problem has received substantial attention in recent years, with methods ranging from wholly unsupervised bottom-up approaches~\cite{grundmann-cvpr2010,xu-eccv2012,galasso-accv2012}, to propagation methods that exploit user input on the first frame~\cite{ren-cvpr2007,tsai-bmvc2010,cipolla-cvpr2010,fathi-bmvc2011,sudheendra-eccv2012,suyog-eccv2014,Wen_2015_CVPR}, to human-in-the-loop methods where a user closely guides the system towards a good segmentation output~\cite{wang-tog2005,li-tog2005,price-iccv2009,bai-2009,Nagaraja_2015_ICCV}.
Successful video segmentation algorithms have potential for significant impact on tasks like  activity and object recognition, video editing, and abnormal event detection.

Despite very good progress in the field, it remains challenging to collect quality video segmentations at a large scale.  The system is expected to segment objects for which it may have no prior model, and the objects may move quickly and change shape or appearance over time---or (often even worse) never move with respect to the background.  To scale up the ability to generate well-segmented data, \emph{human-in-the-loop} methods that leverage minimal human input are appealing~\cite{ren-cvpr2007,tsai-bmvc2010,cipolla-cvpr2010,fathi-bmvc2011,vondrick-nips2011,sudheendra-eccv2012,suyog-eccv2014,Wen_2015_CVPR,wang-tog2005,li-tog2005,price-iccv2009,bai-2009,Nagaraja_2015_ICCV}.  \SDJ{These methods benefit greatly by combining the respective strengths of humans and machines}. They can reserve for the human the more difficult high-level job of identifying a true object, while reserving for the algorithm the more tedious low-level job of propagating that object's boundary over time. \SDJ{This synergistic interaction between humans and computers results in accurate segmentations with minimal huamn effort.}

Critical to the success of an interactive video segmentation algorithm is how the user interacts with the system.  In particular, how should the user indicate the spatial extent of an object of interest in video?  Existing methods largely rely on the tried-and-true interaction modes used for image labeling; namely, the user draws a bounding box or an outline around the object of interest on a given frame, and that region is propagated through the video either indefinitely or until it drifts~\cite{ren-cvpr2007,tsai-bmvc2010,fathi-bmvc2011,sudheendra-eccv2012,suyog-eccv2014,Wen_2015_CVPR}. Furthermore, regardless of the exact input modality, the common assumption is to get the user's input \emph{first}, and then generate a segmentation hypothesis thereafter.  In this sense, in video segmentation propagation, information flows first from the user to the system.



We propose to reverse this standard flow of information.  Our idea is for the system itself to first hypothesize plausible object segmentations in the given frame, and then allow the human user to efficiently and interactively prioritize those hypotheses.  
Such an approach stands to reduce human annotation effort, since the user can use very simple feedback to guide the system to its best hypotheses.

To this end, we introduce \emph{Click Carving}, a novel method that uses point clicks to obtain a foreground object mask for a video frame.  Clicks, largely unexplored for video segmentation, are an attractive input modality due to their ease, speed, and intuitive nature (e.g., with a touch screen the user may simply point a finger).  Our method works as follows.  First, the system precomputes thousands of mask hypotheses based on object proposal regions.  Importantly, those object proposal regions exploit both image coherence cues as well as motion boundaries computed in the video.  Then, the user efficiently navigates to the best hypotheses by clicking on the boundary of the true object and observing the refined top hypotheses.

Essentially, the user's clicks ``carve" away erroneous hypotheses whose boundaries disagree with the clicks.  By continually revising its top rated hypotheses, the system implicitly guides the user where input is most needed next.  After the user is satisfied, or the maximal budget of clicks is exhausted, the system propagates the best mask hypothesis through the video with an existing propagation algorithm.  For videos in the three datasets we tested, only \KG{2-4} clicks are typically required to accurately segment the entire clip. Note that our novel idea is not so much about the ``clicking" interface itself; rather our new ideas center around the idea of simple point supervision as a sufficient cue to perform semi-automatic segmentation and the carving backend that efficiently discerns the most reliable proposals.

Aside from testing our approach with real users, we also develop several simulated user clicking models in order to systematically analyze the relative merits of different clicking strategies.  For e.g., is it more effective to click in the object center, or around its perimeter?  How should multiple clicks be spaced?  Is it advantageous to place clicks in reaction to where the system currently has the greatest errors?  One interesting outcome of our study is that the behavior one might assume as a default---clicking in the object's interior~\cite{Bearman15,Wang201414}---is much less effective than clicking on its boundaries.  We show that boundary clicks are better able to discriminate between good and bad object proposal regions.

The results show that Click Carving strikes an excellent balance of accuracy and human effort. It is faster (requires less annotation interaction) than most existing interactive methods, yet produces better results.  In extensive comparisons with state-of-the-art methods on \KG{three challenging datasets we show that Click Carving outperforms all similarly fast methods, and is competitive or better than those requiring 2 to 12 times the effort.} \SDJ{This large reduction in annotion time by effective use of human interaction naturally leads to large savings in annotation costs. Because of the ease with which our framework can assist even non-experts in making high quality annotations, it has great promise for scaling up video segmentation.  Ultimately such tools are critical for accelerating data collection in several research communities (e.g., computer vision, graphics, medical imaging), where large-scale spatio-temporal annotations are lacking and/or often left to experts.}


\section{Related Work}

\begin{figure*}[t]
  \centering
  \captionsetup{width=\textwidth, font={small}, skip=2pt}
   \begin{tabular}{ccccc}
  \includegraphics[keepaspectratio=true,scale=0.135]{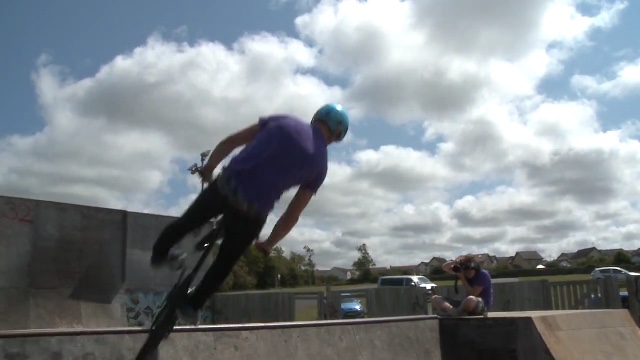} &
  \includegraphics[keepaspectratio=true,scale=0.135]{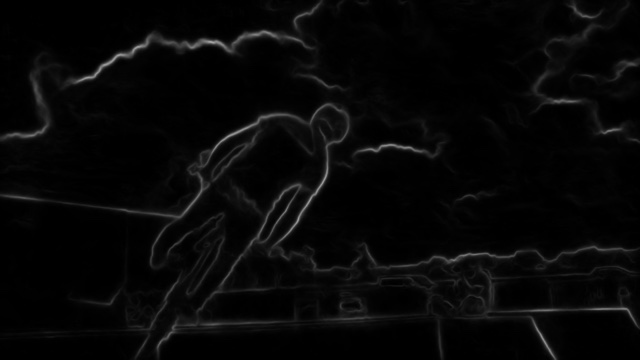} &
  \includegraphics[keepaspectratio=true,scale=0.135]{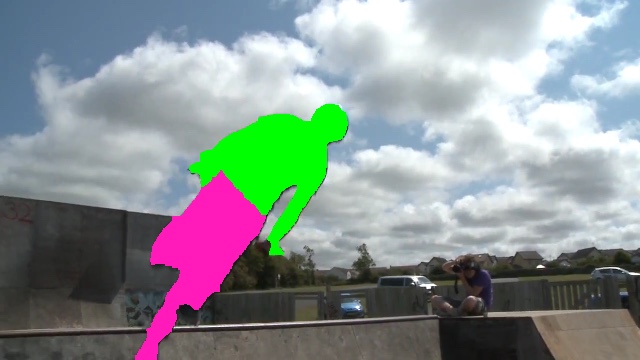} &
  \includegraphics[keepaspectratio=true,scale=0.135]{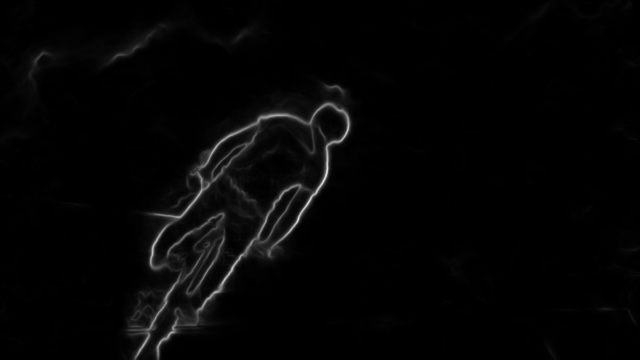} &
  \includegraphics[keepaspectratio=true,scale=0.135]{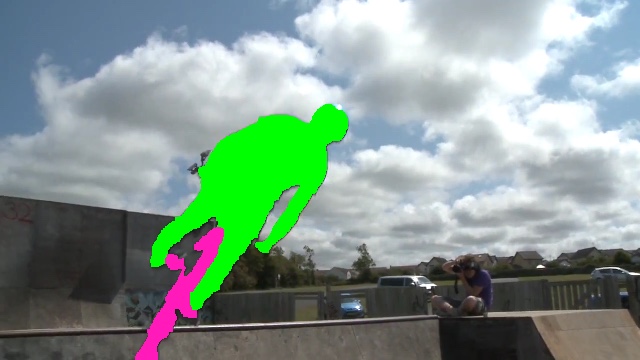} \\
  {\scriptsize (a) Image} & {\scriptsize (b) Static Boundaries} &{\scriptsize (c) Static Proposals} & {\scriptsize (d) Motion Boundaries} & {\scriptsize (e) Motion Proposals} \\

    \end{tabular}
    \caption{Generation of object region proposals using both static and dynamic cues. Best viewed on pdf. }
  \label{fig:preprocess}
\end{figure*}

Unsupervised video segmentation methods use no human input, and typically produce an over-segmentation of the video that is useful for mid-level grouping.  Supervoxel methods find space-time blobs cohesive in color and/or motion~\cite{grundmann-cvpr2010,xu-eccv2012,galasso-accv2012}, while point trajectory approaches  find consistent motion threads beyond optical flow~\cite{brox-eccv2010,lezama-cvpr2011}.  Unlike unsupervised work, we consider interactive video segmentation and our method produces spatio-temporal tubes covering the extent of the complete object.

Other methods extract ``object-like" segments in video~\cite{keysegments,ma-cvpr2012,shah-cvpr2013,rehg-iccv2013,ferrari-iccv2013}, typically by learning the category-independent properties of good regions, and employing some form of tracking.  Related are the methods that generate a large number of bounding box or region \emph{proposals}~\cite{Wu_2015_CVPR,Fragkiadaki_2015_CVPR,Yu_2015_CVPR,oneata}, an idea originating in image segmentation~\cite{cpmc,APBMM2014}.  The idea is to maintain high recall for the sake of downstream processing.  As such, these methods typically produce many segmentation hypotheses, \KG{100s to 1000s for today's popular datasets}.  To adapt them for the object segmentation problem would require human inspection to select the best one, which is non-trivial once the video contains more than a handful.  Our approach makes use of object proposals, but our idea to prioritize them with Click Carving is entirely new.  \KG{We are the first to propose using proposals for the sake of speeding up interactive segmentation, whether for images or videos.}


Semi-automatic video segmentation methods accept manually labeled frame(s) as input, and propagate the annotation to the remaining video clip~\cite{ren-cvpr2007,tsai-bmvc2010,cipolla-cvpr2010,fathi-bmvc2011,sudheendra-eccv2012,suyog-eccv2014,Wen_2015_CVPR}.  Often the model consists of an MRF over (super)pixels or supervoxels, with both spatial and temporal connections. Some systems
   actively guide the user how to annotate~\cite{fathi-bmvc2011,vondrick-nips2011,sudheendra-eccv2012,karasevRS14}.
 All the prior methods assume that initialization starts with the user, and all use traditional modes of user input (bounding boxes, scribbles, or outlines).  In contrast, we explore the utility of clicks for video segmentation, and we propose a novel, interactive way to perform system-initiated initialization. We show our approach achieves comparable performance to drawing complete outlines, but with much less annotation effort.  Following our novel user interaction stage, we rely on an existing propagation method and make no novelty claims about how the propagation stage itself is done.


\SDJ{Human-in-the-loop systems have proved to very useful in diverse computer vision tasks such as training object detectors~\cite{tropel}, counting objects~\cite{jellybean} etc. Interactive video segmentation methods also leverage user input, but unlike the above propagation methods, the user is always in the loop and engages in a back and forth until the video is adequately segmented~\cite{wang-tog2005,bai-2009,Nagaraja_2015_ICCV}. } In all existing methods, the user guides the system to generate a segmentation hypothesis, and then iteratively corrects mistakes by providing more guidance.  In contrast, Click Carving pre-generates thousands of possible hypotheses and then employs user guidance to efficiently filter high quality segmentations from them.  \KG{Our approach could potentially be used in conjunction with many of these systems as well, to reduce the interaction effort.}  Compared to propagation methods, the interactive methods usually have the advantage of greater precision, but at the disadvantages of greater human effort and less amenability to crowdsourcing.

Only limited work explores click supervision for image and video annotation.  Clicks on objects in images can remove ambiguity to help train a CNN for semantic segmentation from weakly labeled images~\cite{Bearman15}, or to spot object instances in images for dataset collection~\cite{LinECCV14coco}.  Clicks on patches are used to obtain ground truth material types in~\cite{bell15minc}.

We are aware of only two prior efforts in video segmentation using clicks, and their usage is quite different than ours.  In one, a click and drag user interaction is used to segment objects~\cite{PFS2015}.  A small region is first selected with a click, then dragged to traverse up in the hierarchy until the segmentation does not bleed out of the object of interest. Our user-interaction is much simpler (jut a few mouse clicks or taps on the touchscreen) and our boundary clicks are discriminative enough to quickly filter good segmentations.  In the other, the TouchCut system uses a single touch to segment the object using level-set techniques~\cite{Wang201414}. However, the evaluation is focused on image segmentation, with only limited results on video; our approach outperforms it.



\section{Approach}
\label{sec:approach}

We now present our approach.

\subsection{Generating video foreground proposals}\label{sec:proposals}

\KG{Existing propagation-based video segmentation methods rely on human input
(a bounding box, contour, or scribble) at the onset to generate results.}
The key idea behind our Click Carving approach is to
flip this process. \KG{Instead of the human annotator providing a foreground
region from scratch, the system generates many plausible segmentation mask hypotheses and
the annotator efficiently navigates to the best ones with point clicks.}

Specifically, we use state-of-the-art region proposal generation algorithms
to generate 1000s of possible foreground segmentations for the first video
frame.\footnote{\KG{For clarity of presentation, we describe the process as
always propagating from the first annotated frame.  However, the system can
be initialized from arbitrary frames.}} Region proposal methods aim to obtain high recall at the cost of low precision.
Even though this guarantees that at least a few of these segmentations will
be of good quality, it is difficult to filter out the best ones automatically
with existing techniques.


To generate accurate region proposals in videos, we use the multiscale
combinatorial grouping (MCG) algorithm~\cite{APBMM2014} with both static and
motion boundaries.  The original algorithm uses image boundaries to obtain a
hierarchical segmentation, followed by a grouping procedure to obtain
region-based foreground object proposals. The video datasets that we use in
this work have both static and moving objects. We observed that due to
factors like motion blur etc., static image boundaries are not very reliable
in many cases. On the other hand, optical flow provides a strong cue about
the objects contours while the object is in motion. Hence we also use motion
boundaries~\cite{weinzaepfel:hal-01142653} to generate per-frame motion
region proposals using MCG.  The two sources are complementary in nature: for
static objects, the per-frame region proposals obtained using static
boundaries will be more accurate, and vice versa.

Figure~\ref{fig:preprocess} illustrates this with an example. Both the person
and bike (Figure \ref{fig:preprocess}a) are in motion.  As a result, we get
weaker static boundaries (Figure \ref{fig:preprocess}b). Figure
\ref{fig:preprocess}c shows the best static proposal for each object; the
proposal quality for the bike is very poor. On the other hand, the motion
boundaries (Figure \ref{fig:preprocess}d) are much stronger and result in
very accurate proposals for both the person and the bike (Figure
\ref{fig:preprocess}e).

In summary, given a video frame, we generate the set
of foreground region proposals ($\mathcal{M}$)  for it by taking the union
between the static region proposals ($\mathcal{M}_{static}$) and motion
region proposals ($\mathcal{M}_{motion}$), i.e., $\mathcal{M} =
\{\mathcal{M}_{static} \cup \mathcal{M}_{motion}\}$. On average we generate a
total of about 2000 proposals per frame, resulting in a very high overall
recall. \SDJ{(The Mean Average Best Overlap score (MABO) is 78.3 on the three datasets that we use. 
This is computed by selecting the proposal with highest overlap score in each frame and taking a dataset-wide average).}  \KG{ In what follows, we explain how Click
Carving allows a user to efficiently navigate to the best proposal among
these thousands of candidates.}

\begin{figure*}[t]
	\centering
	\scriptsize
	\begin{tabular}{cx{3mm}cccccc}
		{\bf User Click} & & {\bf ContourMap} & 
		\multicolumn{5}{c}{\bf Top-5 ranked proposals} \\
		\includegraphics[keepaspectratio=true,scale=0.12]{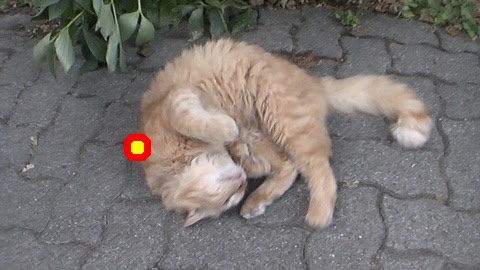} &
		&
		\includegraphics[keepaspectratio=true,scale=0.12]{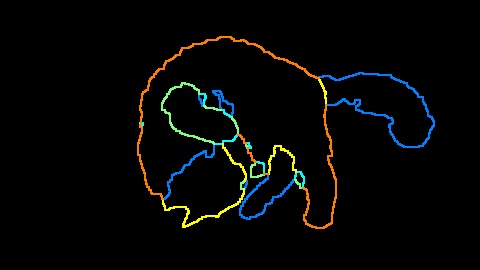} &
		\includegraphics[keepaspectratio=true,scale=0.12]{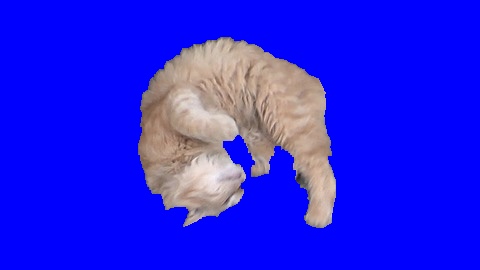} &
		\includegraphics[keepaspectratio=true,scale=0.12]{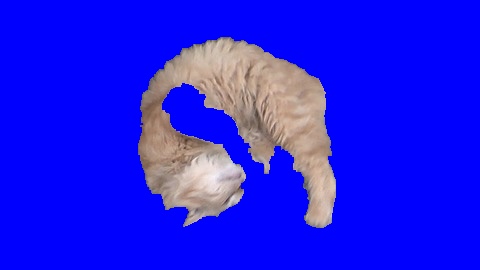} &
		\includegraphics[keepaspectratio=true,scale=0.12]{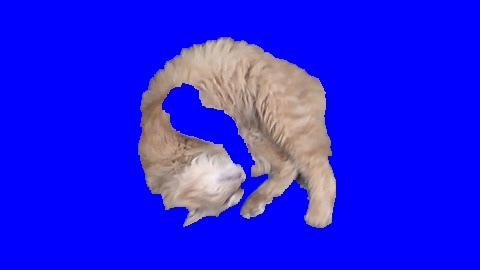} &
		\includegraphics[keepaspectratio=true,scale=0.12]{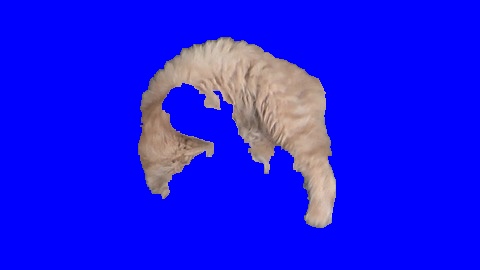} &
		\includegraphics[keepaspectratio=true,scale=0.12]{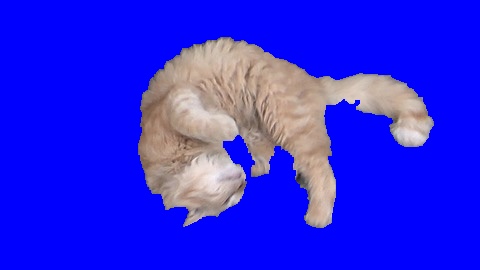} \\

		\includegraphics[keepaspectratio=true,scale=0.12]{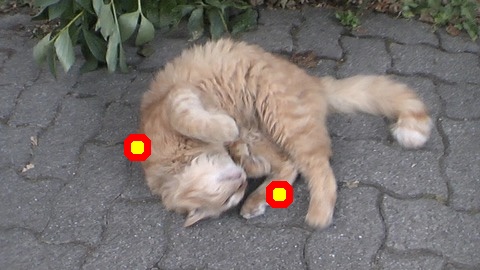} &
		&
		\includegraphics[keepaspectratio=true,scale=0.12]{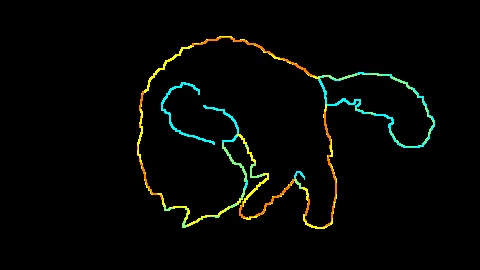} &
		\includegraphics[keepaspectratio=true,scale=0.12]{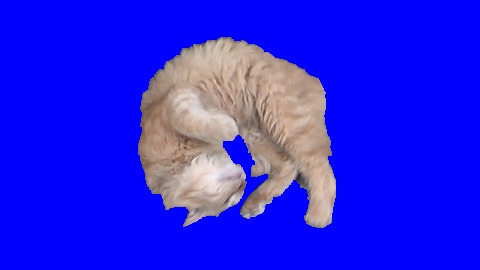} &
		\includegraphics[keepaspectratio=true,scale=0.12]{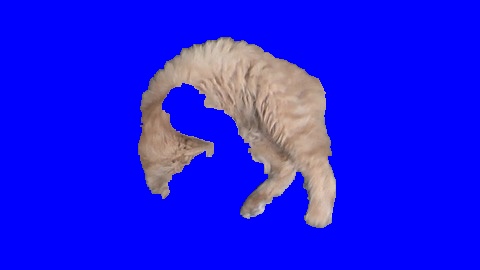} &
		\includegraphics[keepaspectratio=true,scale=0.12]{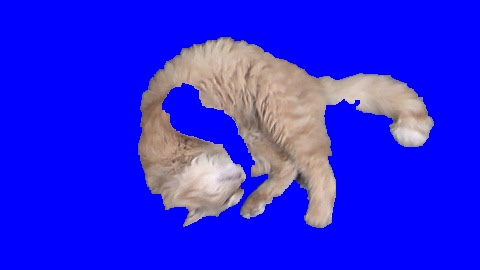} &
		\includegraphics[keepaspectratio=true,scale=0.12]{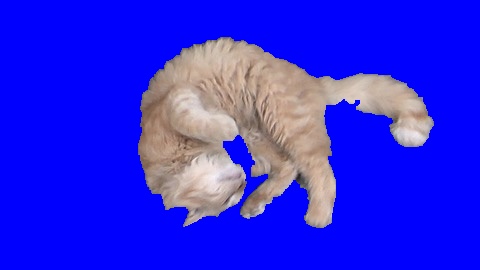} &
		\includegraphics[keepaspectratio=true,scale=0.12]{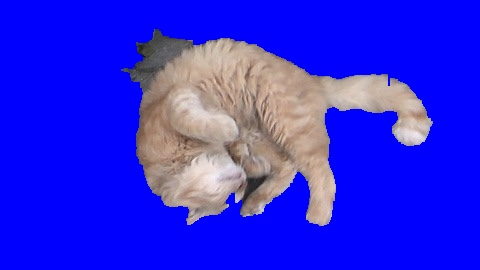} \\
		
		\includegraphics[keepaspectratio=true,scale=0.11]{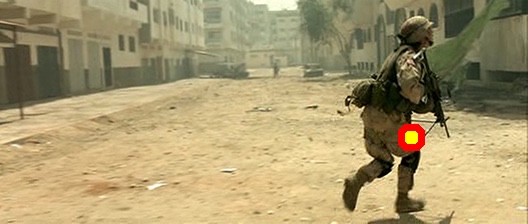} &
		&
		\includegraphics[keepaspectratio=true,scale=0.11]{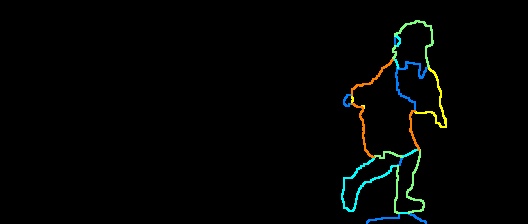} &
		\includegraphics[keepaspectratio=true,scale=0.11]{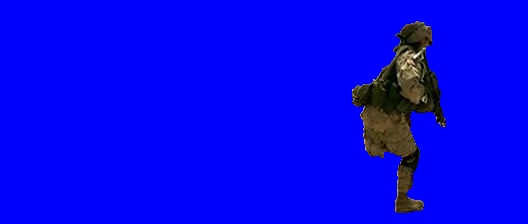} &
		\includegraphics[keepaspectratio=true,scale=0.11]{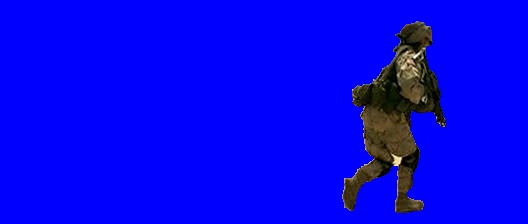} &
		\includegraphics[keepaspectratio=true,scale=0.11]{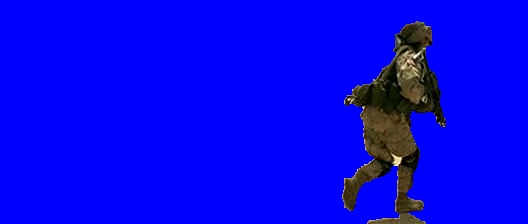} &
		\includegraphics[keepaspectratio=true,scale=0.11]{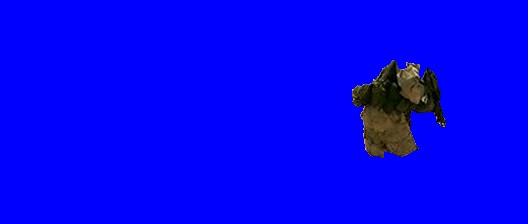} &
		\includegraphics[keepaspectratio=true,scale=0.11]{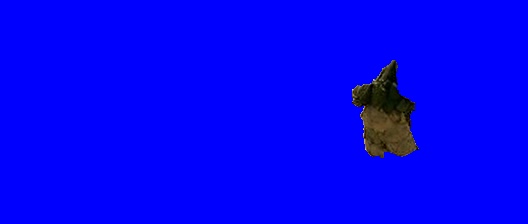} \\
		
		\includegraphics[keepaspectratio=true,scale=0.11]{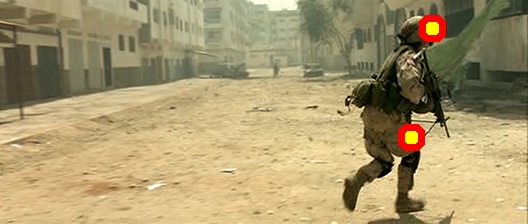} &
		&
		\includegraphics[keepaspectratio=true,scale=0.11]{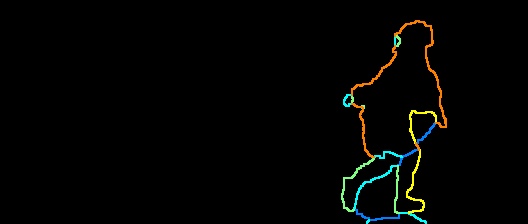} &
		\includegraphics[keepaspectratio=true,scale=0.11]{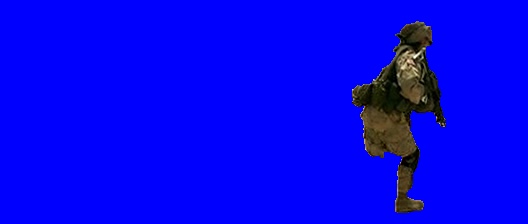} &
		\includegraphics[keepaspectratio=true,scale=0.11]{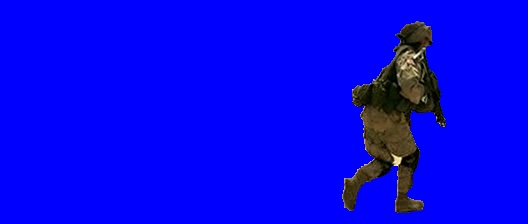} &
		\includegraphics[keepaspectratio=true,scale=0.11]{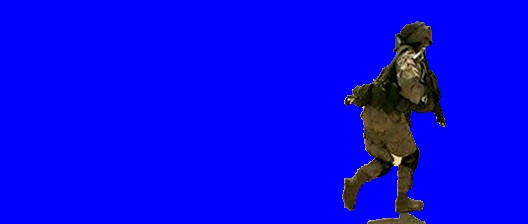} &
		\includegraphics[keepaspectratio=true,scale=0.11]{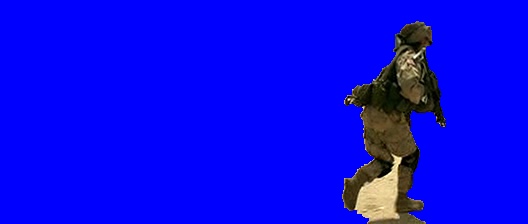} &
		\includegraphics[keepaspectratio=true,scale=0.11]{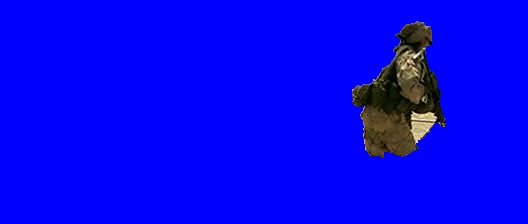} \\
		
		\includegraphics[keepaspectratio=true,scale=0.11]{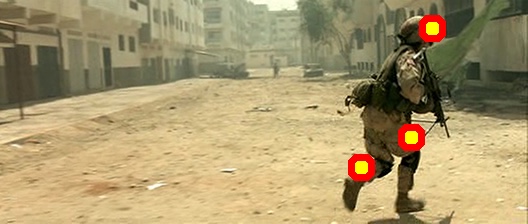} &
		&
		\includegraphics[keepaspectratio=true,scale=0.11]{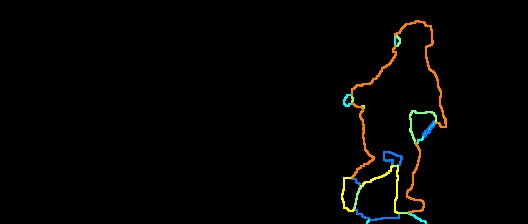} &
		\includegraphics[keepaspectratio=true,scale=0.11]{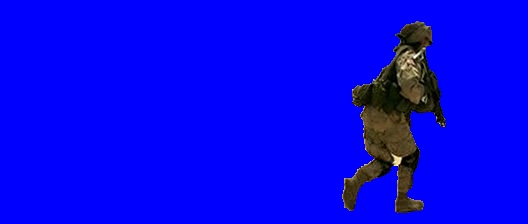} &
		\includegraphics[keepaspectratio=true,scale=0.11]{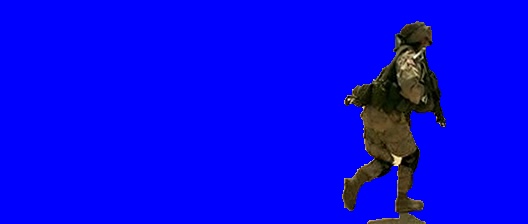} &
		\includegraphics[keepaspectratio=true,scale=0.11]{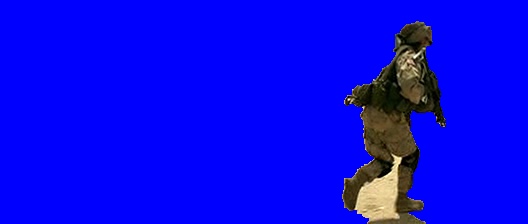} &
		\includegraphics[keepaspectratio=true,scale=0.11]{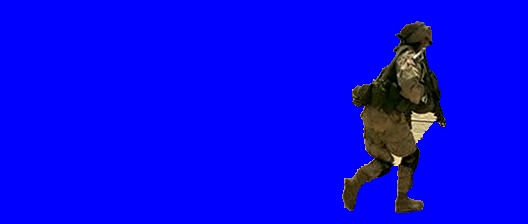} &
		\includegraphics[keepaspectratio=true,scale=0.11]{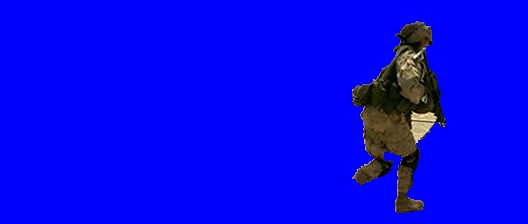} \\

	\end{tabular}
	\caption{Click Carving based foreground segmentation.  Best viewed on pdf.  See text for details.}
	\label{fig:approach}
	\vspace{4pt}
\end{figure*}

\subsection{Click Carving for discovering an object mask}

The region proposal step yields a large set of segmentation hypotheses
(1000s), out of which only a few are very accurate object segmentations.
\KG{A naive approach that asks an annotator to manually scan through all
proposals is both tedious and inefficient.}  We now explain how our Click
Carving algorithm effectively and very quickly identifies the quality
segmentations. We show that within a few clicks, it is possible to obtain a
very high quality segmentation of the desired object of interest.  \KG{We
stress that while Click Carving assists in getting the mask for a single
frame, it is closely tied to the video source due to the motion-based
proposals.}

At a high level, our Click Carving algorithm converts the user clicks into
votes cast for the underlying region proposals. The user initiates the
algorithm by clicking somewhere on the boundary of the object of interest.
This click casts a vote for all the proposals whose boundaries also
\KG{(nearly)} intersect with the user click. Using these votes, the
underlying region proposals are re-ranked and the user is presented with the top-$k$ proposals having the highest votes.

\KG{This process of clicking and re-ranking iterates.  At any time,} the user
can choose any of the top-$k$ as the final segmentation if he/she is
satisfied, or he/she can continue to re-rank by clicking and casting more
votes.

More specifically, we characterize each proposal, $\mathcal{M}_j \in
\mathcal{M}$ with the following four components
($\mathcal{M}_j^m,\mathcal{M}_j^e,\mathcal{M}_j^s,\mathcal{M}_j^v$):

\begin{itemize}

\item Segmentation mask ($\mathcal{M}_j^m$): This quantity represents the
actual region segmentation mask obtained from the MCG region proposal
algorithm.

\item Contour mask ($\mathcal{M}_j^e$): Our algorithm requires the user to
click on the object boundaries, which as we will show later is much more
discriminative than clicking on interior points and results in a much faster
filtering of good segmentations. To infer the votes on the boundaries, we
convert the segmentation mask $\mathcal{M}_j^m$ into a contour mask. This
contour mask only contains the boundary pixels from $\mathcal{M}_j^m$. For
error tolerance, we dilate the boundary mask by 5 pixels on either side. This
reduces the sensitivity of the exact user click location, \KG{which need not
coincide exactly with the mask boundary.}

\item Objectness score ($\mathcal{M}_j^s$): We use the objectness score from
the MCG algorithm~\cite{APBMM2014} to break ties if multiple region proposals get
the same number of votes.  \SJ{This score reflects the likelihood of a given region to be an accurate object segmentation.}

\item User votes ($\mathcal{M}_j^v$): This quantity represents the total
number of user votes received by a particular proposal at any given time.
\KG{It is initialized to 0.}

\end{itemize}

As a first step, we begin by computing a lookup table which allows us to
efficiently account for the votes cast for each proposal by the user. Let $n$
be the total number of pixels in a given image and $m$ be the total number of
region proposals generated for that image. We define and precompute a lookup
table $\mathcal{T} \in \{0,1\}^{n \times m}$ as follows:

\begin{equation}
\mathcal{T}(i,j) =
\left\{
  \begin{array}{ll}
      \vspace{5pt}
   1 &  \text{ if } \mathcal{M}_j^e(i)=1  \\
   0  &  \text{ otherwise},
  \end{array}
\right. \label{eq:lut} \end{equation} where $i$ denotes a particular pixel
and $j$ denotes a particular region proposal.

When the user clicks at a particular pixel location $c$, the weights for each
of the region proposal are updated as follows: \begin{equation}
\mathcal{M}_j^v = \mathcal{M}_j^v + \mathcal{T}(c,j).
  \label{eq:vots}
\end{equation}

The updated set of votes is used to re-rank all the region proposals. The
proposals with equal votes are ranked in the order of their objectness
scores. This interactive re-ranking procedure continues until the user is
satisfied with any of the top-$k$ proposals and chooses that as the final
segmentation. \KG{In our implementation, $k$ is set such that $k$ copies of
the image, one proposal on each, fit easily on one screen
 ($k=9$).}

Figure~\ref{fig:approach} illustrates our user interface and explains this
process with two examples. We show the user interaction on the leftmost
column.  Red circles denote clicks. The ``ContourMap" column shows the average contour map of the top-5 ranked proposals after the user click. \KG{Here the colors are a heat-map
coding of the number of votes for a boundary fragment.}
Remaining columns show the top-5 ranked proposals.

The top two rows show an example frame from a ``cat" video in the
iVideoSeg~\cite{Nagaraja_2015_ICCV} dataset. The user\footnote{\SJ{We discuss
our user study in the next section.}} places the first click on the
left side of the object (top left image). We see that the resulting top
ranked proposals (5 foreground images in top row) align well to the current
user click, \KG{meaning they all contain a boundary near the click point.}
The average contour map of these top ranked proposals, informs the user
about areas that have been carved well already (red lines) and which areas
may need more attention (blue lines, \KG{or contours on the true object that
remain uncolored}). The user observes that \KG{most} current top-$k$
segmentations are missing the cat's right leg and decides to place the next
click there (second row, leftmost image). The next ranking of the proposals
brings up segmentations which cover the entire object accurately.

In the next example, we consider a frame from the ``soldier" video in the
Segtrack-v2 dataset~\cite{rehg-iccv2013}.   The user decides to place a click
on the right side of the object (third row, leftmost image). This click
itself retrieves a very good segmentation for the soldier. However, to
explore further, the user continues by making more clicks. Each new
constraint eliminates the bad proposals from the previous step, and after
just 3 clicks, all the top-ranked proposals are of good quality.

\subsection{User clicking strategies}
\label{sec:users}
To quantitatively evaluate Click Carving, we employ both real human
annotators and simulated users with different clicking strategies. \KG{We
design a series of clicking strategies to simulate, each of which represents
a hypothesis for how a user might efficiently convey which object boundaries
remain missing in the top proposals.  While real users are arguably the best
way to judge final impact of our system (and so we use them), the simulated user models are
complementary.  They allow us to run extensive trials and to see at scale
which strategies are most effective.} Simulated human users have also been
studied in interactive segmentation for brush stroke
placement~\cite{kohli-ijcv}.

We categorize the user models into three groups: human annotators, boundary
clickers, and interior clickers.
\vspace{2pt}

{\bf Human annotators:} We conduct a user study to analyze the
performance of our method by recruiting \KG{3} human annotators to work on
each image. \SJ{The 3 annotators included a computer vision student and 2 non-expert users.} The human annotators were encouraged to click on object boundaries,
while observing the current best segmentations. \KG{They were also given some time to familiarize themselves with the interface, before starting the actual experiments.}  They had a
choice to stop by choosing one of the segmentations among the top ranked ones
or continue clicking to explore further. A maximum budget of 10 clicks was
used to limit the total annotation time. The target object was indicated to
them before starting the experiment. In the case of multiple objects, each
object was chosen as the target object in a sequential manner. We recorded
the number of clicks, time spent, and the best object mask chosen by the user
during each segmentation. The user corresponding to the median number of
clicks is used for our quantitative evaluation.

\vspace{2pt}
{\bf Boundary clickers:} We design three simulated users which operate by
clicking on object boundaries. To simulate these artificial users, we make
use of the ground-truth segmentation mask of the target object. Equidistant
points are sampled from the ground truth object contour to define object
boundaries. Each simulated boundary clicker starts from the same initial
point. We use principal component analysis (PCA) on the ground truth shape to
find the axis of maximum shape variation. We consider a ray from the centroid
of the object mask along the direction of this principal axis. The furthest
point on the object boundary where this ray intersects is chosen as the
starting point. The three boundary clickers that we design differ in how they
make subsequent clicks from this starting point. They are: \vspace{5pt}\\  (a)
\emph{\textbf{Uniform clicker:}} To obtain uniformly spaced clicks, we divide
the total number of boundary points  by the maximum click budget to obtain a
fixed distance interval $d$. Starting from the initial point and walking
along the boundary, a click is made every $d$ points apart from the previous
click location. \vspace{5pt}\\ (b) \emph{\textbf{Submod clicker:}} The uniform user has a
high level of redundancy, since it clicks at locations which are still close
to the previous clicks; hence the gain in information between two consecutive
clicks might be small. Next we design a boundary clicker that tries to impact
the maximum boundary region with each subsequent clicks. This is done by
placing the click at a boundary point which is furthest away from its nearest
user click among all boundary points. This resembles the sub-modular subset
selection problem\KG{~\cite{krause07nectar}}, where one tries to maximize the set
coverage while choosing a subset. We employ a greedy algorithm to find the
next best point.\vspace{5pt}\\ (c) \emph{\textbf{Active clicker:}} The previous two
methods only looked at the ground truth segmentation to devise a click
strategy, without taking into account the segmentation performance after each
click is added. Our active clicking strategy takes into account the current
best segmentation among the top-$k$ (vs.~the ground truth) and uses that to
make the next click decision. It is
 similar in design to the Submod user, except that it skips those
boundary points which have already been labeled correctly by the top-ranked
proposal. We find that this active simulated user comes the closest in
mimicking the actual human annotators (see results for details). 

\vspace{2pt}
{\bf Interior clickers:} A novel insight of our method is the discriminative
nature of boundary clicks. In contrast, default behavior and \KG{previous
user models~\cite{Wang201414,Bearman15}} assumes a click in the interior of the
object is well-suited.  To examine this contrast empirically, our final
simulated user clicks on interior object points.  To simulate interior
clicks, we uniformly sample object pixel locations from the entire ground
truth segmentation mask (up to the maximum click budget) and then
sequentially place clicks on the object of interest.

We analyze the impact of the click strategies in our results section.

\subsection{Propagating the mask through the video}
 \SJ{Having discovered a good object mask using Click Carving in the initial frame, the next step is to propagate this segmentation to all other frames in the video. We use the foreground propagation method of~\cite{suyog-eccv2014} as our segmentation method primarily due to good performance and efficiency, using code from the authors. We also tried other methods like~\cite{Wen_2015_CVPR} but found~\cite{suyog-eccv2014} to be most scalable for large experiments. In its original form the method requires a human drawn object outline in the initial frame. We instead initialize the method using the region proposal which was selected using Click Carving. This initial mask is then propagated to the entire video to obtain the final segmentation. We computed the supervoxels required by~\cite{suyog-eccv2014} using~\cite{grundmann-cvpr2010} and use the default parameter settings.
}

\section{Results}\label{sec:results}

\subsection{Datasets and metrics} 

We evaluate on \KG{3} publicly available datasets: Segtrack-v2~\cite{rehg-iccv2013}, VSB100~\cite{Sundberg,NB13} and iVideoSeg~\cite{Nagaraja_2015_ICCV}. For evaluating segmentation accuracy we use the standard intersection-over-union (IoU) overlap metric between the predicted and ground-truth segmentations. A brief overview of the datasets:
\begin{itemize}
\item {\bf SegTrack v2 ~\cite{rehg-iccv2013}:} the most common benchmark to evaluate video object segmentation. It consists of 14 videos with a total of 24 objects and 976 frames. Challenges include appearance changes, large deformation, motion blur etc. Pixel-wise ground truth (GT) masks are provided for every object in all frames.
\vspace{2pt}
\item {\bf Berkeley Video Segmentation Benchmark (VSB100) ~\cite{Sundberg,NB13}:} consists of 100 HD sequences with multiple objects in each video. We use the ``train" subset of this dataset in our experiments, for a total of 39 videos and \SJ{4397} frames. This is a very challenging dataset; interacting objects and small object sizes make it difficult to segment and propagate. We use the GT annotations of multiple foreground objects provided by~\cite{PFS2015} on every 20th frame.

\vspace{2pt}
\item {\bf iVideoSeg ~\cite{Nagaraja_2015_ICCV}:} This new dataset consists of 24 videos from 4 different categories (car, chair, cat, dog). Some videos have viewpoint changes and others have large object motions. GT masks are available for 137 of all 11,882 frames.
\end{itemize}

\subsection{Methods for comparison} 

We compare with state-of-the art \SJ{methods~\cite{ferrari-iccv2013,Nagaraja_2015_ICCV,Wen_2015_CVPR,keysegments,Wang201414,grundmann-cvpr2010,rehg-iccv2013,suyog-eccv2014,godec11a}} and our own baselines. Below we group them into 6 groups based on the amount of human annotation effort, i.e., the interaction time between the human and algorithm. In some cases, a human simply initializes the algorithm, while in others the human is in the loop always.

{\bf (1) Unsupervised:} We use the state-of-the-art method of \textbf{\cite{ferrari-iccv2013}}, which produces a single \SJ{region} segmentation result per video with zero human involvement.


\vspace{3pt}
{\bf (2) Multiple segmentation:} Most existing unsupervised methods produce multiple segmentations to achieve high recall. We consider both {\bf 1) Static object proposals (BestStaticProp):} where the best per frame region proposal (out of approx 2000 proposals per frame) is chosen as the final segmentation for that frame {\bf 2) Spatio-temporal proposals~\cite{rehg-iccv2013,keysegments,grundmann-cvpr2010}:} These methods produce multiple spatio-temporal region tracks as segmentation hypotheses. To simulate a human picking the desired segmentation from the hypotheses, we use the dataset ground truth to select the most overlapping hypothesis.  We use the duration of the video to estimate interaction time. \KG{This is a lower bound on cost,} since the annotator has to at least watch the clip once to select the best segmentation.  \KG{For the static proposals, we multiply the number of frames by 2.4 seconds, the time required to provide one click~\cite{Bearman15}.}
\vspace{3pt}

{\bf (3) Scribble-based:} We consider two existing methods: {\bf 1) JOTS~\cite{Wen_2015_CVPR}:} the first frame is interactively segmented using scribbles and GrabCut. The segmentation result is than propagated to the entire video. We use the timing data from the detailed study by~\cite{McGuinness2010434}, who find 
it takes a human on average 66.43 seconds per image to obtain a good segmentation with scribbles.  \KG{{\bf 2) iVideoSeg~\cite{Nagaraja_2015_ICCV}:} This is a recently proposed state-of-the-art technique that uses scribbles to interactively label point trajectories. These labels are then used to segment the object of interest. We use the timing data kindly shared by the authors.}
\vspace{3pt}

{\bf (4) \KG{Object outline} propagation:} the human outlines the object completely to initialize the propagation algorithm (typically in the first frame), which then propagates to the entire video. 
\KG{Here we use the same method for propagation~\cite{suyog-eccv2014} as in our approach.}
Timing data from~\cite{suyog-iccv2013,LinECCV14coco} indicate it typically takes 54-79 seconds to manually outline an object; we use the more optimistic 54 seconds for this baseline.  
\vspace{3pt}

{\bf (5) Bounding box:} Rather than segment the object, the annotator draws a tight bounding box around it. 
The baseline {\bf BBox-GrabCut} uses that box to obtain a segmentation for the video as follows.  We learn a Gaussian Mixture Model (GMM) based appearance model for foreground and background pixels according to the box, then apply them in a standard spatio-temporal MRF defined over pixels. The unaries are derived from the learnt GMM model and contrast-sensitive spatial and temporal potentials are used for smoothness.
\vspace{3pt}

{\bf (6) 1-Click based:} We also consider baselines which perform video segmentation with a single user click. {\bf 1) TouchCut~\cite{Wang201414}} the only prior work using clicks for video segmentation.  {\bf 2) Click-GrabCut:} This is similar to BBox-GrabCut except that we take a small region around the click to learn the foreground model. The background model is learnt from a small area around image boundaries. {\bf 3) Click-STProp:} To propagate the impact of a user click to the entire video volume, we use the spatio-temporal proposals from~\cite{oneata}. We do this by selecting all proposals which enclose the click inside them. Fg and bg appearance models are learnt using the selected proposals and refined using a spatio-temporal MRF. We use the timing data from~\cite{Bearman15}, which reports that a human takes about 2.4 seconds to place a single click on the object of interest.

\begin{table*}[t]
	\centering
	\scriptsize
	\captionsetup{width=\textwidth, font={scriptsize}, skip=2pt}
	\begin{tabular}{|c|c|c|c|c|c|c|c|c||c|}
		\hline
		& & Objectness & Interior & BBox & Uniform & Submod & Active & Human & BestProp  \\ \hline
		\multirow{3}{*}{\bf Segtrack-v2 }  & Clicks & 0 & 6.29 & 2 & 4.46 & 3.83 & 3.34 & {\bf 2.46}  & -  \\ \cline{2-10}
		& Time (sec) & 0 & 23.95 & 7 & 16.98 & 14.58 & 12.72 & {\bf 9.37} & -  \\ \cline{2-10}
		& IoU & 42.36 & 52.79 & 67.51 & 75.8 & 76.76 & 76.24 & {\bf 78.77} & 80.74  \\ \hline
		\hline
		\multirow{3}{*}{\bf VSB100 }  & Clicks & 0 & 7.05 & 2 & 5.34 & 5.28 & 5.23 & {\bf 4.35} & -  \\ \cline{2-10}
		& Time (sec) & 0 & 30.11 & 7 & 22.81 & 22.55 & 22.33 & {\bf 18.58} & -  \\ \cline{2-10}
		& IoU & 28.45 & 46.98 & 58.98 & 64.2 & 65.67 & 66.91 & {\bf 69.63} & 72.82  \\ \hline
		\hline
		\multirow{3}{*}{\bf iVideoSeg }  & Clicks & 0 & 5.02 & 2 & 3.84 & 3.29 & 3.15 & {\bf 2.84} & -  \\ \cline{2-10}
		& Time (sec) & 0 & 19.86 & 7 & 15.20 & 13.02 & 12.47 & {\bf 11.24} & -  \\ \cline{2-10}
		& IoU & 50.69 & 72.54 & 68.04 & 77.57 & 77.84 & {\bf 78.65} & 78.24 & 81.34  \\ \hline

	\end{tabular}
	\caption{Click-carving proposal selection quality for real users (Human), the different user click models (Interior, Uniform, Submod, Active), and a \KG{BBox baseline}.  With an average of 2-4 clicks to carve the proposal boundaries, users attain IoU accuracies very close to the upper bound (BestProp).  Objectness, Interior clicks and the BBox baseline are substantially weaker. IoU measures segmentation overlap with the ground truth; perfect overlap is 100.
	}
	\label{results-click-carving}
\end{table*}

\begin{figure*}[t]
	\captionsetup{width=\textwidth, font={scriptsize}, skip=2pt}
	\begin{tabular}{c|c}
		{\includegraphics[keepaspectratio=true,scale=0.20]{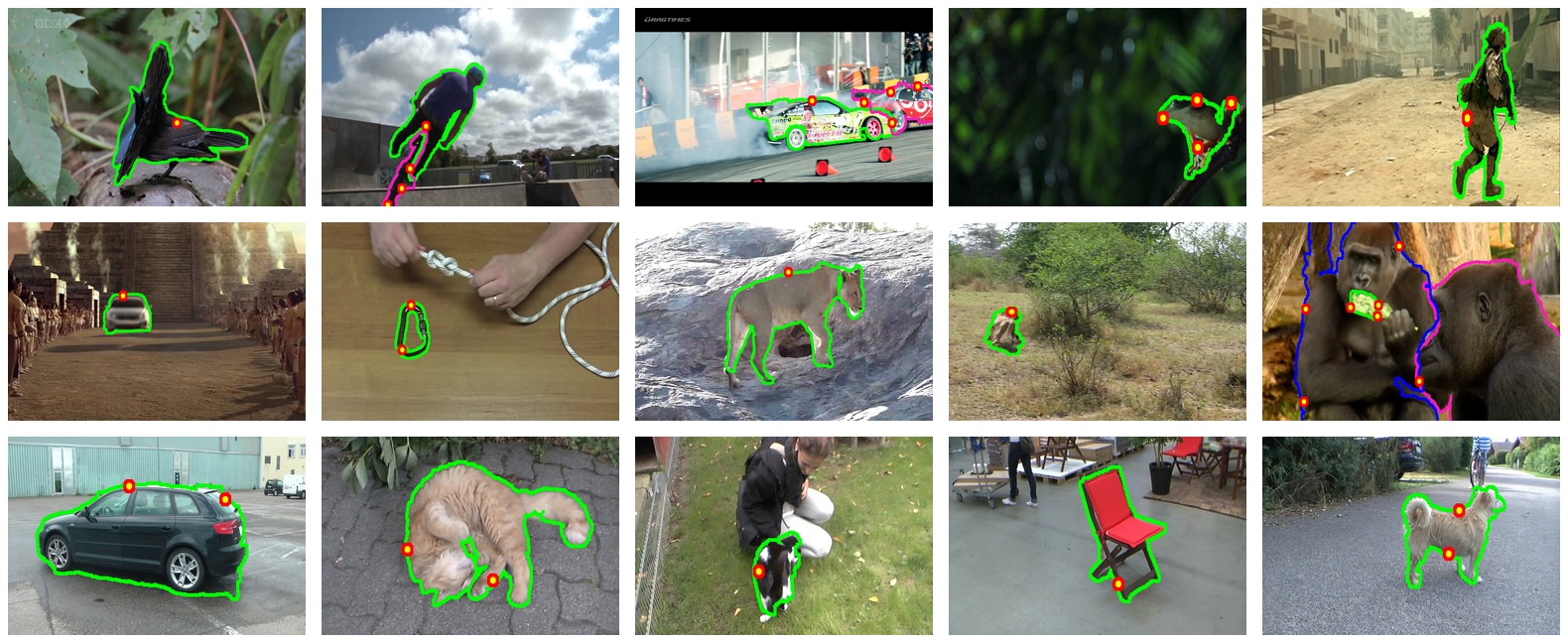}} &
		{\includegraphics[keepaspectratio=true,scale=0.27]{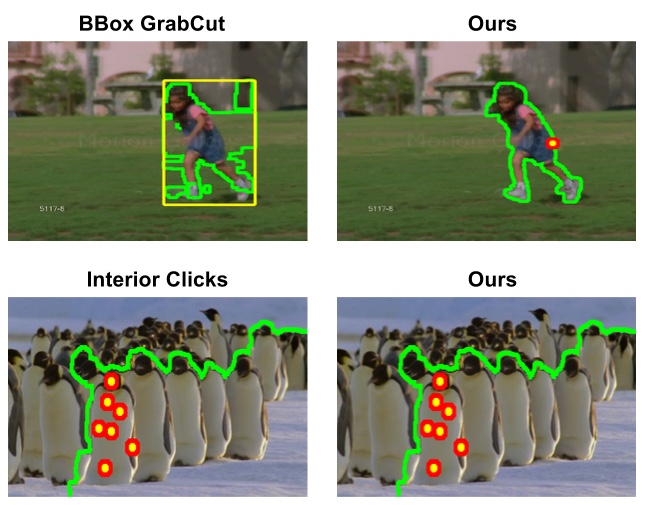}}  \\
		Visual results for Click-Carving & Visual comparisons with baselines 
	\end{tabular}
	\caption{{\bf Left:} Qualitative results for Click Carving. The yellow-red dots show the clicks made by human annotators. The best selected segmentation boundaries are overlayed on the image (green). {\bf Right:} Comparisons with baselines: The top example shows the segmentation we obtain with a single click as opposed to applying GrabCut segmentation with a tight bounding box. Bottom example shows the discriminative power of clicking on boundaries by comparing it with a baseline which clicks in the interior regions. Best viewed on pdf. }
	\label{fig:qual_result}
\end{figure*}

\subsection{Experiments}
We first test the accuracy/speed tradeoff in terms of locating the best available proposal, and compare the simulated user models.  Then we present comparisons against all the existing methods and baselines.

\vspace{5pt}
\noindent {\bf Click Carving for region proposal selection:} We first present the performance of Click Carving for interactively locating the best region proposal for the object of interest. We do this for the first frame in all videos. 
In all experiments, we set the total click budget to be a maximum of 10 clicks per object. For simulated users, clicks are placed sequentially depending on its design, until a proposal which is within 5\% overlap of the best proposal is ranked in the top-$k$ or the click budget is exhausted. For the human user study, the user stops when they decide that they found a good segmentation within the top-$k$ ranked proposals or have exhausted the click budget.

Table~\ref{results-click-carving} shows the results for all datasets and compares the performance with all simulated users.  We compare both in terms of the number of clicks and time required and also how close they get to the best proposal available in the pool of $\sim$2000 (BestProp). As expected, in all cases real users achieve the best segmentation performance and require far fewer clicks than all simulated users to achieve it. Our simulated Active user, which takes into account the current state of the segmentation, comes closest to matching the human's performance. Also, we see clicking uniformly on the object boundaries requires more clicks on average than the Active and Submodular users, which try to impact the largest object area with each subsequent click. The objectness baseline, which first ranks all the proposals using objectness scores and picks the best proposal among top-$k$ ($k$=9) performs the worst. This shows that user interaction is key to picking good quality proposals among 1000s of candidates.

All users that operate by clicking on boundaries (Human, Uniform, Submod and Active), come very close to choosing the best proposal in most cases. In contrast, clicking on the interior points requires substantially more clicks---often double the time; more importantly, the best segmentation it obtains is much worse in quality than the best possible segmentation.  
This supports our hypothesis that clicking on boundaries is much more discriminative in separating good proposals from the bad ones. Whereas a matching between an object proposal contour and a boundary click will rarely be accidental, several bad proposals may have the interior click point lie within them.

In fact, selecting the best proposal using an enclosing bounding box around the true object (BBox) is more effective than clicking on interior points. This is likely because a tight bounding box can eliminate a large number of proposals that extend outside its boundaries. On the other hand, an interior click cannot restrict the selected proposals to the ones which align well to the object boundaries. Our method outperforms the bounding box selection by a large margin, showing the efficacy of our approach.

On Segtrack-v2 and iVideoSeg, Click Carving requires less than 3 clicks on average to obtain a high quality segmentation. For the most challenging dataset, VSB100, we obtain good results with an average of 4.35 clicks. This shows the potential of our method to collect large amounts of segmentation data economically. The timing data reveals the efficiency and scalability of our method.  Below we show how this translates to advantages for complete video segmentation.

Figure \ref{fig:qual_result} (left) show qualitative results for Click Carving. In many cases (e.g., lions, soldier, cat), only a single click is sufficient to obtain a high quality segmentation. Several challenging instances like the cat (bottom row) and the lion (middle row), are segmented very accurately with a single click. These objects would otherwise require a large amount of human interaction to obtain good segmentation \KG{(say using a GrabCut like approach).}
More clicks are typically needed when multiple objects are close-by or interacting with each other. Still, we observe that in many cases only a small number of clicks on each object results in good segmentations. For example, in the car video (top row), only 5 clicks are required to obtain final segmentations for both objects.

\SJ{Figure \ref{fig:qual_result} (right) highlights the key strengths of our method over two baselines. In the top example, we see that GrabCut segmentation applied even with a very tight bounding box fails to segment the object. On the other hand, even with a single click, our proposed approach produces very accurate segmentation. The example on the bottom shows the importance of clicking on boundaries. Clicking on the interior fails to retrieve a good proposal, because several bad proposals also contain those interior clicks. But our boundary clicks, which are highly discriminative, retrieve the best proposal quickly.}

\begin{table*}[t]
	\centering
	\tiny
	\captionsetup{width=\textwidth, font={scriptsize}, skip=2pt}
	\begin{tabular}{|c|c|x{10mm}|x{10mm}|x{10mm}|x{8mm}|x{9mm}|x{8mm}|x{10mm}|x{10mm}|x{8mm}|x{6mm}|x{8mm}|}
		\hline
		&  & \textbf{Unsup.} & \multicolumn{4}{c|}{\textbf{Multiple Segmentations}} & \textbf{Scribbles} & \textbf{\KG{Outline}} & {\textbf{Bounding Box}} & \multicolumn{3}{c|}{\textbf{Click Based}} \\ \hline
		
		& \textbf{\#Frames} & \textbf{\cite{ferrari-iccv2013}} & \textbf{\cite{grundmann-cvpr2010}} & \textbf{\cite{keysegments}} & \textbf{\cite{rehg-iccv2013}} & \textbf{BestStaticProp} & \textbf{\cite{Wen_2015_CVPR}} & \textbf{\cite{suyog-eccv2014}} & \textbf{BBox-GrabCut} & \textbf{Click-GrabCut} & \textbf{Click-STProp} & {\textbf{Ours}}  \\ \hline
		birdfall2 & 30 & 32.28 & 57.40 & 49.00 & 62.50 & 72.00 & 78.70 & 63.15 & 2.59  & 2.12 & 59.21 & 62.32  (1) \\ \hline
		bird of paradise & 97 & 81.83 & 86.80 & 92.20 & 94.00 & 93.11 & 93.00 & 91.59 & 35.46  & 28.04 & 86.24 & 89.90  (1) \\ \hline
		bmx - person & 36 & 51.81 & 39.20 & 87.40 & 85.40 & 86.20 & 88.90 & 82.74 & 14.99  & 7.45 & 77.81 & 81.14  (1) \\ \hline
		bmx - cycle & 36 & 21.71 & 32.50 & 38.60 & 24.90 & 64.27 & 5.70 & 2.95 & 4.73  & 2.52 & 13.21 & 2.15  (3) \\ \hline
		cheetah - deer & 29 & 40.32 & 18.80 & 44.50 & 37.30 & 59.61 & 66.10 & 33.15 & 7.46  & 5.30 & 20.24 & 29.87  (5) \\ \hline
		cheetah - cheetah & 29 & 16.53 & 24.40 & 11.70 & 40.90 & 62.51 & 35.30 & 26.96 & 9.47  & 6.32 & 24.42 & 19.96  (1) \\ \hline
		drift-1 & 74 & 52.35 & 55.20 & 63.70 & 74.80 & 85.50 & 67.30 & 69.15 & 18.57  & 14.72 & 54.45 & 68.51  (2) \\ \hline
		drift-2 & 74 & 33.18 & 27.20 & 30.10 & 60.60 & 78.92 & 63.70 & 52.49 & 16.59  & 15.13 & 51.97 & 49.91  (3) \\ \hline
		frog & 279 & 54.13 & 67.10 & 0.00 & 72.80 & 78.10 & 56.30 & 69.69 & 49.65  & 28.95 & 64.91 & 69.12  (4) \\ \hline
		girl & 21 & 54.90 & 31.90 & 87.70 & 89.20 & 72.06 & 84.60 & 67.38 & 28.26  & 18.21 & 63.43 & 66.56  (1) \\ \hline
		hummingbird-1 & 29 & 8.97 & 13.70 & 46.30 & 54.40 & 77.55 & 58.30 & 58.63 & 24.02  & 19.64 & 42.12 & 44.24  (1) \\ \hline
		hummingbird-2 & 29 & 32.10 & 25.20 & 74.00 & 72.30 & 83.48 & 50.70 & 56.24 & 59.05  & 28.94 & 44.67 & 39.95  (1) \\ \hline
		monkey & 31 & 64.20 & 61.90 & 79.00 & 84.80 & 85.87 & 86.00 & 73.86 & 39.93  & 28.67 & 69.14 & 72.60  (2) \\ \hline
		monkeydog - monkey & 71 & 72.33 & 68.30 & 74.30 & 71.30 & 77.58 & 82.20 & 74.32 & 14.22  & 12.86 & 67.80 & 71.12  (1) \\ \hline
		monkeydog - dog & 71 & 0.02 & 18.80 & 4.90 & 18.90 & 57.19 & 21.10 & 75.47 & 3.30  & 2.45 & 35.10 & 65.21  (4) \\ \hline
		parachute & 51 & 76.32 & 69.10 & 96.30 & 93.40 & 90.40 & 94.40 & 87.78 & 92.92  & 85.80 & 87.78 & 86.22  (1) \\ \hline
		penguin 1 & 42 & 5.09 & 72.00 & 12.60 & 51.50 & 79.98 & 94.20 & 92.09 & 16.13  & 9.87 & 14.56 & 77.41  (2) \\ \hline
		penguin 2 & 42 & 2.16 & 80.70 & 11.30 & 76.50 & 87.85 & 91.80 & 79.70 & 17.09  & 14.65 & 16.34 & 76.45  (3) \\ \hline
		penguin 3 & 42 & 1.86 & 75.20 & 11.30 & 75.20 & 84.17 & 91.90 & 91.62 & 13.44  & 12.34 & 14.24 & 81.43  (3) \\ \hline
		penguin 4 & 42 & 2.31 & 80.60 & 7.70 & 57.80 & 82.31 & 90.30 & 76.92 & 9.44  & 9.53 & 18.19 & 73.26  (3) \\ \hline
		penguin 5 & 42 & 9.95 & 62.70 & 4.20 & 66.70 & 77.48 & 76.30 & 77.12 & 15.87  & 9.54 & 14.32 & 76.54  (7) \\ \hline
		penguin 6 & 42 & 18.88 & 75.50 & 8.50 & 50.20 & 83.46 & 88.70 & 80.65 & 10.79  & 10.23 & 21.34 & 80.21  (3) \\ \hline
		solider & 32 & 39.77 & 66.50 & 66.60 & 83.80 & 80.30 & 81.10 & 72.10 & 35.38  & 21.20 & 79.80 & 71.29  (1) \\ \hline
		worm & 154 & 72.79 & 34.70 & 84.40 & 82.80 & 83.73 & 79.30 & 72.99 & 13.52  & 8.94 & 67.10 & 72.20  (5) \\ \hline
		\hline
		\textbf{\KG{Average Accuracy}} &  - & 35.24 & 51.89 & 45.26 & 65.92 & 78.48 & 71.91 & 67.86 & 23.04  & 16.81 & 46.18 & 63.65 \\ \hline
		\textbf{Annotation Effort} &  & - & 336.6 tracks & 10.6 tracks & 60 tracks & 120k proposals & 1 frame & 1 frame &  2 clicks & 1 click & 1 click & 2.46 clicks  \\ \hline
		\textbf{Annotation Time (sec)} &  & 0 & 673.2 & 21.2 & 120 & \SJ{142.5} & 66.43 & 54 & 7  & 2.4 & 2.4 & 9.37  \\ \hline

	\end{tabular}
	\caption{Video segmentation accuracy (IoU) on Segtrack-v2 (per-video). The last column shows our result with real human users. Numbers in parens are the \# of clicks required by our method. The bottom two rows summarize the amount of human annotation effort required to obtain the corresponding segmentation performance, for all methods.  Our approach leads to the best trade-off between video segmentation accuracy and human annotation effort.
	}
	\vspace*{0.1in}
	\label{results-segtrack}
\end{table*}

\begin{table*}[t]
	\centering
	\tiny
	\captionsetup{width=0.6\textwidth, font={scriptsize}, skip=2pt}
	\begin{tabular}{|c|x{15mm}|x{15mm}|x{14mm}|x{8mm}|x{8mm}|x{8mm}|}
		
		\hline
		&   \textbf{Unsup.}  & \textbf{\KG{Outline}} & \textbf{Bounding Box} & \multicolumn{3}{c|}{\textbf{Click Based}} \\ \hline
		
		&  \textbf{\cite{ferrari-iccv2013}}   & \textbf{\cite{suyog-eccv2014}} & \textbf{BBox-GrabCut} &  \textbf{Click-GrabCut} & \textbf{Click-STProp} & {\textbf{Ours}} \\ \hline
		\textbf{\KG{Avg. Accuracy}}  & 17.79  & 61.43 & 14.74   & 11.14 & 26.76 & 56.15 \\ \hline
		\textbf{Annot. Effort}   & -  & 1 frame & 2 clicks  & 1 click & 1 click & 4.35 clicks  \\ \hline
		\textbf{Annot. Time (sec)}   &  0  & \SJ{54} & 7  & 2.4 & 2.4 & 18.58 \\ \hline
	\end{tabular}
	\caption{Video segmentation accuracy (IoU) on all 39 videos in VSB100, format as in Table~\ref{results-segtrack}.  Our approach provides the best trade-off between video segmentation accuracy and human annotation effort. 
	}
	\vspace*{0.1in}
	\label{results-vsb}
\end{table*}

\begin{figure*}[t]
  \centering
  \captionsetup{width=\textwidth, font={scriptsize}, skip=2pt}
   \begin{tabular}{ccc}
  \includegraphics[keepaspectratio=true,scale=0.30]{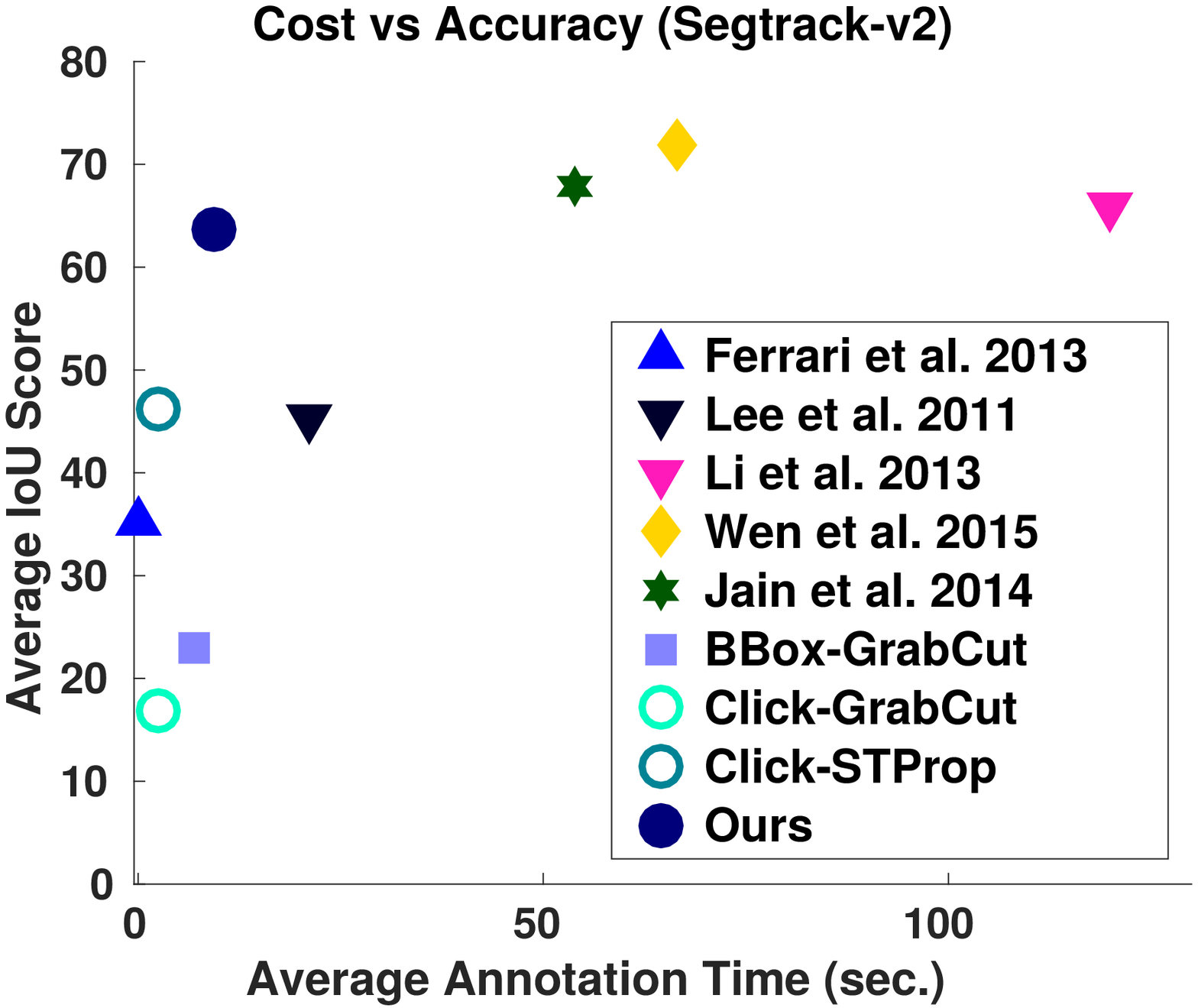} &
  \includegraphics[keepaspectratio=true,scale=0.30]{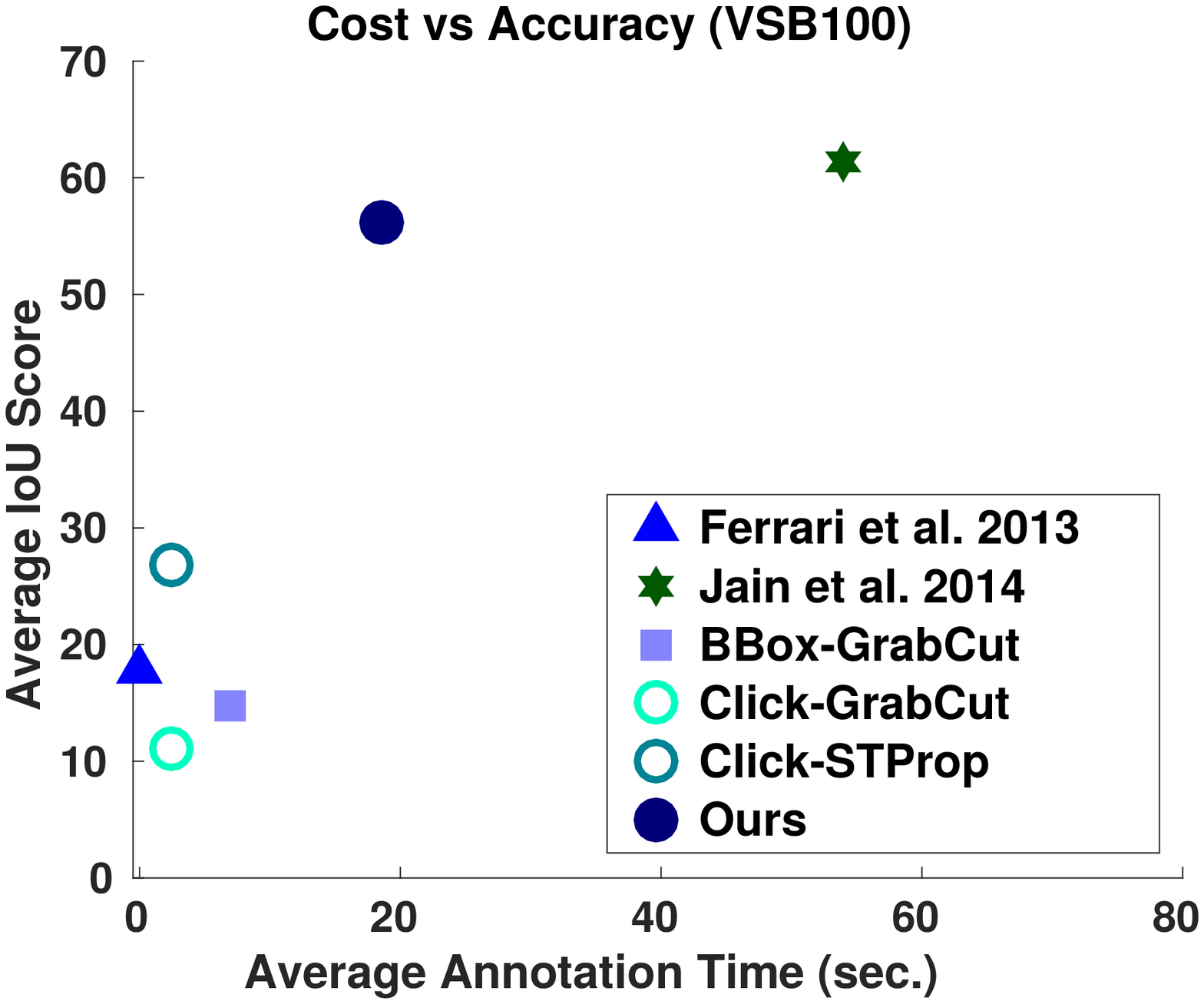} &
    \includegraphics[keepaspectratio=true,scale=0.30]{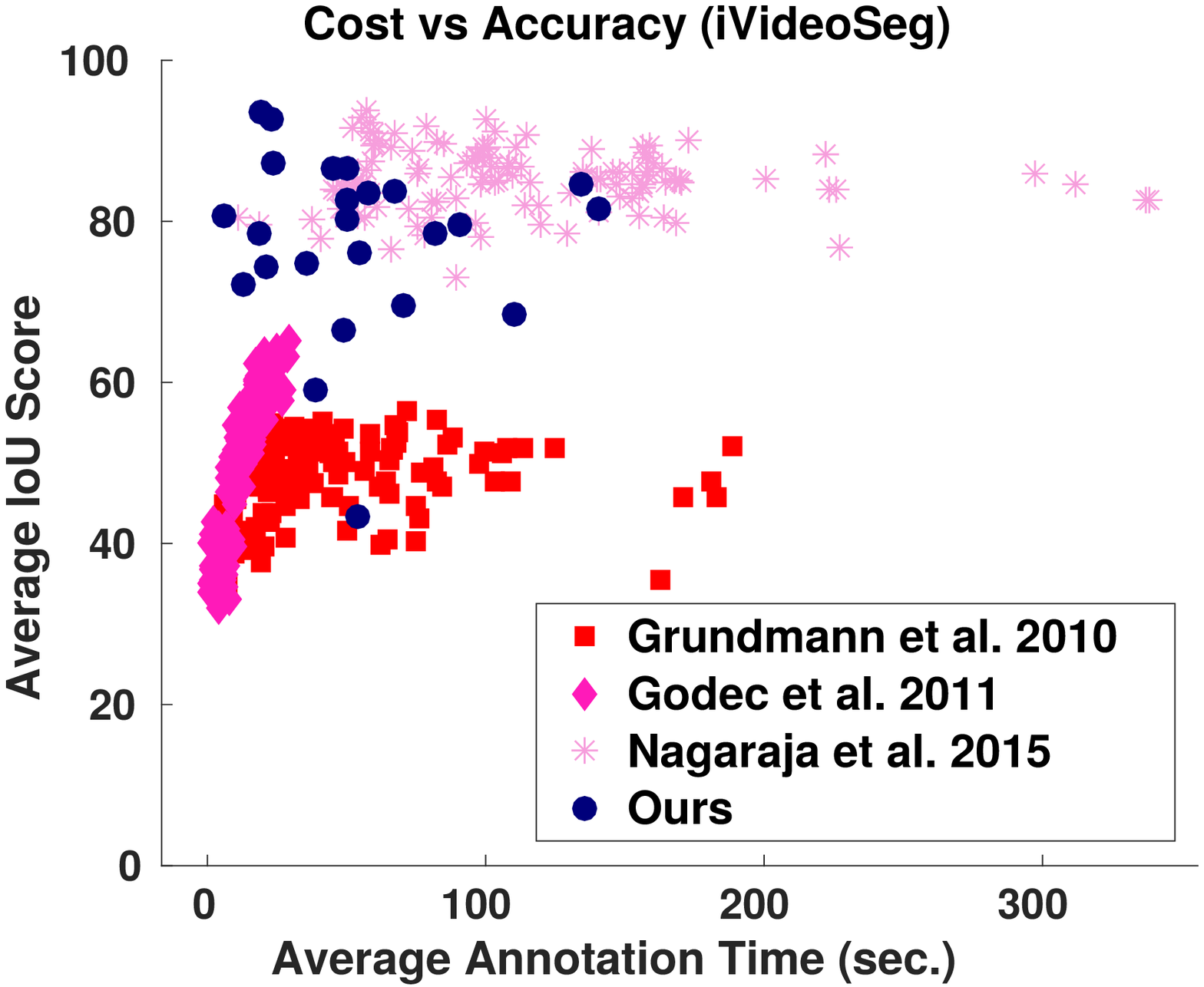}
    \end{tabular}
    \caption{\SJ{Cost vs accuracy on Segtrack (left), VSB100 (center), and iVideoSeg (right). Our Click Carving based video propagation results in similar accuracy as state-of-the-art metods, but it does so with much less human effort.  Click Carving offers the best trade-off between cost and accuracy. Best viewed on pdf.} }
  \label{fig:cost_accuracy}
\end{figure*}

Next we discuss the results for video segmentation, where we propagate the results of Click Carving to the remaining frames in the video.

\vspace{5pt}
\noindent  {\bf Video segmentation propagation on Segtrack-v2:} Table \ref{results-segtrack} shows the results on Segtrack-v2. We compare using the standard intersection-over-union (IoU) metric with a total of \SJ{10} methods which use varying amounts of human supervision. The unsupervised algorithm~\cite{ferrari-iccv2013} that uses no human input results in the lower accuracy.  Among the approaches which produce multiple segmentations, BestStaticProp and~\cite{rehg-iccv2013} have the best accuracy. This is expected because these methods are designed for having high recall, but it requires much more effort to sift through the multiple hypotheses to pick the best one as the final segmentation. For example, it is prohibitively expensive to go through 2000 segmentations for each frame to get to the accuracy level of BestStaticProp. The method of~\cite{rehg-iccv2013} produces much fewer segmentations, but still requires 12x more time than our method to achieve comparable performance.

The scribble based method~\cite{Wen_2015_CVPR} achieves the best overall accuracy on this dataset, but is 6 times more expensive than our method. The propagation method of~\cite{suyog-eccv2014}, which we also use as our propagation engine, sees an increase of 4\% in accuracy when propagated from human-labeled object outlines. On the other hand, our method which is initialized from slightly imperfect, but much quicker to obtain object boundaries achieves comparable performance. 
Using computer generated segmentations coupled with our Click Carving interactive selection algorithm is sufficient to obtain high performance.

Moving on to the methods that require less human supervision, i.e., bounding boxes and clicks, we see that Click Carving continues to hold advantages. In particular, BBox-GrabCut and Click-GrabCut result in poor performance, indicating that more nuanced propagation methods are needed than just relying on appearance-based segmentation alone. Click-STProp, which obtains a spatial prior by propagating the impact of a single click to the entire video volume, results in much better performance than solely appearance based methods. However, our method, which first translates clicks into accurate per-frame segmentation before propagating them, yields a \KG{17\%} gain (\KG{37\% relative gain}).

All these trends show that our method offers the best trade-off between segmentation performance and annotation time. Figure~\ref{fig:cost_accuracy} (left), visually depicts this trade-off. All methods which result in better segmentation accuracy than ours need substantially more human effort. Even then the gap in the performance in relatively small. On the flip side, the methods which require less annotation effort than us also result in a significant degradation in segmentation performance.

\vspace{5pt}
\noindent {\bf Video segmentation propagation on VSB100:} Next, we test on VSB100. This is an even more challenging dataset and very few existing methods have reported foreground propagation results on it. Since this dataset includes several videos that contain multiple interacting objects in challenging conditions, Click Carving tends to require more clicks (4.35 on average).  Our method again outperforms all baselines which require less human effort and results in comparable performance with~\cite{suyog-eccv2014}, but at a much lower cost. Figure ~\ref{fig:cost_accuracy} (center) again reflects this trend.

\vspace{5pt}
\noindent {\bf Video segmentation propagation on iVideoSeg:} We also compare our method on the recently proposed iVideoSeg dataset~\cite{Nagaraja_2015_ICCV}. We compare with 3 methods ~\cite{grundmann-cvpr2010,godec11a,Nagaraja_2015_ICCV} out of which ~\cite{Nagaraja_2015_ICCV} is the current state-of-the-art method for interactive foreground segmentation in videos. We use the timing information provided by the authors~\cite{Nagaraja_2015_ICCV}. We compare the performance of our method on all 24 videos in the dataset (300-1000 frames per video) using the real user annotation times. Figure~\ref{fig:cost_accuracy} (right) shows the results. For all methods, each data point on the plot shows time vs. accuracy for a particular video at a particular iteration. 

The methods of~\cite{Nagaraja_2015_ICCV,grundmann-cvpr2010,godec11a} run for multiple iterations i.e. a human provides annotation on several frames, observes the results and repeats until he/she is satisfied. This requires a human to evaluate the current video segmentation result and decide if more annotation is required. The authors provided timing and accuracy data for 4-5 iterations on each video. In contrast our method does one-shot selection instead of iterative refinement. Our method pre-selects the frames on which to request human annotation (every 100th frame in this case). For each selected frame, we ask a human annotator to use our Click Carving method to find the best region proposal while recording their timing. The total time for the video is simply the sum of time taken for each selected frame. The video segmentation propagation is re-initialized whenever a new labeled frame is available. 

We outperform both~\cite{grundmann-cvpr2010,godec11a} by a considerable margin. When compared with~\cite{Nagaraja_2015_ICCV}, our method achieves similar segmentation accuracy but with less than half the total annotation time. On average over all 24 videos, ~\cite{Nagaraja_2015_ICCV} takes 110.05 seconds to achieve an IoU score of 80.04. In comparison our method only takes 54.35 seconds to reach an IoU score of 77.68.

\vspace{5pt}
\noindent {\bf Comparison with TouchCut:} To our knowledge TouchCut~\cite{Wang201414} is the only prior work which utilizes clicks for video segmentation. In that work, the user places a click somewhere on the object, then a level-sets technique transforms the click to an object contour. This transformed contour is then propagated to the remaining frames. Very few experimental results about video segmentation are discussed in the paper, and code is not available.  Therefore, we are only able to compare with TouchCut on the 3 Segtrack videos reported in their paper. Table~\ref{results-touchcut} shows the result. When initialized with a single click, our method outperforms TouchCut in 2 out of 3 videos. With 1 more click, we perform better in all 3 videos.

\begin{table}[h]
\centering
\scriptsize
\captionsetup{width=0.45\textwidth, font={scriptsize}, skip=2pt}
\begin{tabular}{|c|c|c|c|}
\hline
   & \textbf{TouchCut} & \textbf{Ours (1-click)} & \textbf{Ours (2-clicks)} \\ \hline
birdfall2 & 248 & 213 & 187 \\ \hline
girl & 1691 & 2213 & 1541 \\ \hline
parachute & 228 & 225 & 198 \\ \hline

\end{tabular}
\caption{Comparison with TouchCut~\cite{Wang201414} \KG{in terms of pixel error (lower is better)}.}
\label{results-touchcut}
\end{table}

\begin{figure*}[t]
	\centering
	\captionsetup{width=0.9\textwidth, font={scriptsize}, skip=2pt}
	\begin{tabular}{c}
		\includegraphics[keepaspectratio=true,scale=0.25]{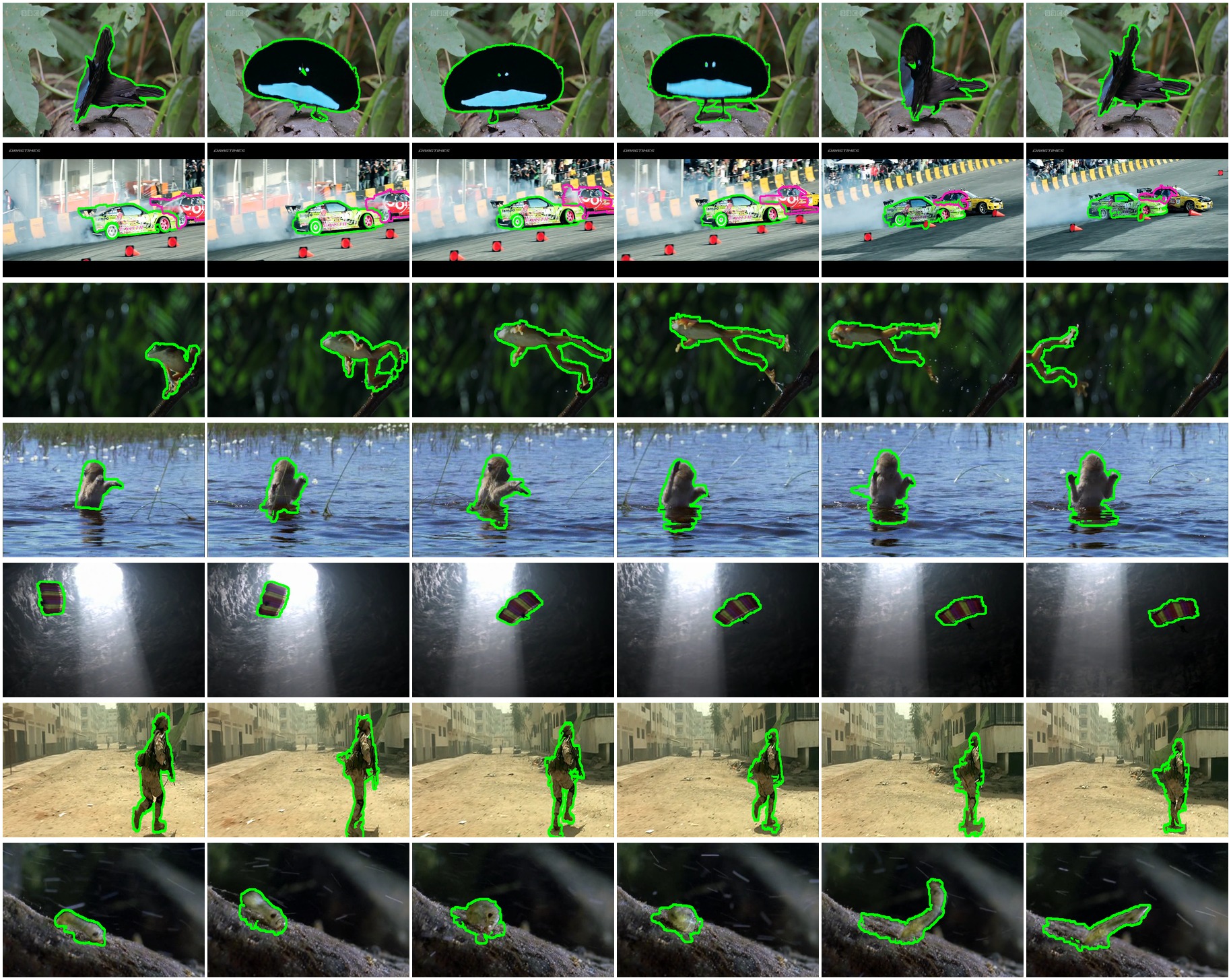}
	\end{tabular}
	\caption{Qualitative results for video segmentation on Segtrack-v2 dataset: The results using the propagation method of~\cite{suyog-eccv2014} initialized from the segmentation in the left-most image. This initialization is obtained using our Click Carving method with static and motion-based proposals. Best viewed on pdf. }
	\label{fig:videoseg_result_segtrack}
\end{figure*}

\begin{figure*}[t]
	\centering
	\captionsetup{width=0.9\textwidth, font={scriptsize}, skip=2pt}
	\begin{tabular}{c}
		\includegraphics[keepaspectratio=true,scale=0.25]{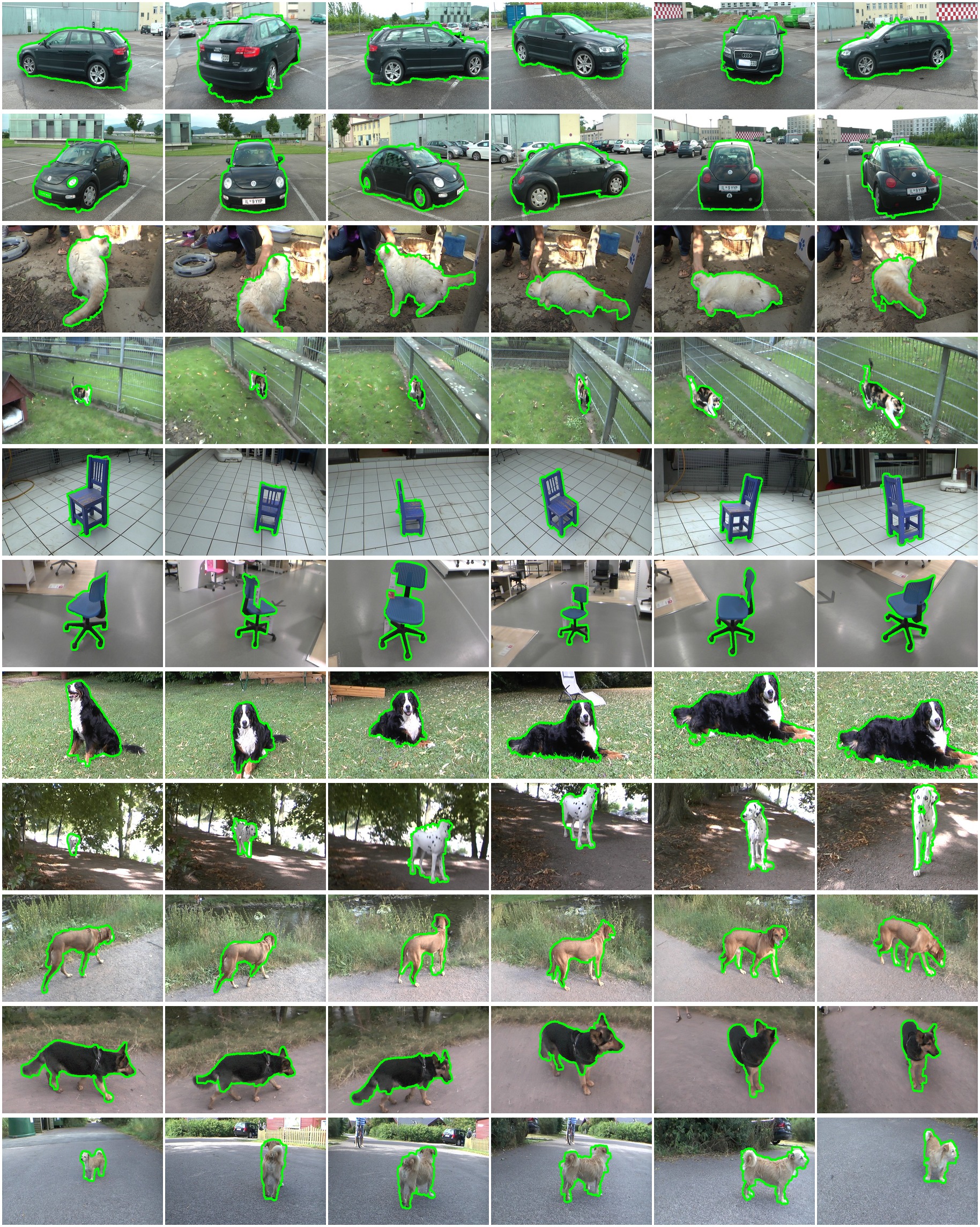}
	\end{tabular}
	\caption{Qualitative results for video segmentation on iVideoSeg dataset: The results using the propagation method of~\cite{suyog-eccv2014} initialized from the segmentation in the left-most image. This initialization is obtained using our Click Carving method with static and motion-based proposals. Best viewed on pdf. }
	\label{fig:videoseg_result_ivideoseg}
\end{figure*}

\begin{figure*}[t]
	\centering
	\captionsetup{width=0.9\textwidth, font={scriptsize}, skip=2pt}
	\begin{tabular}{c}
		\includegraphics[keepaspectratio=true,scale=0.25]{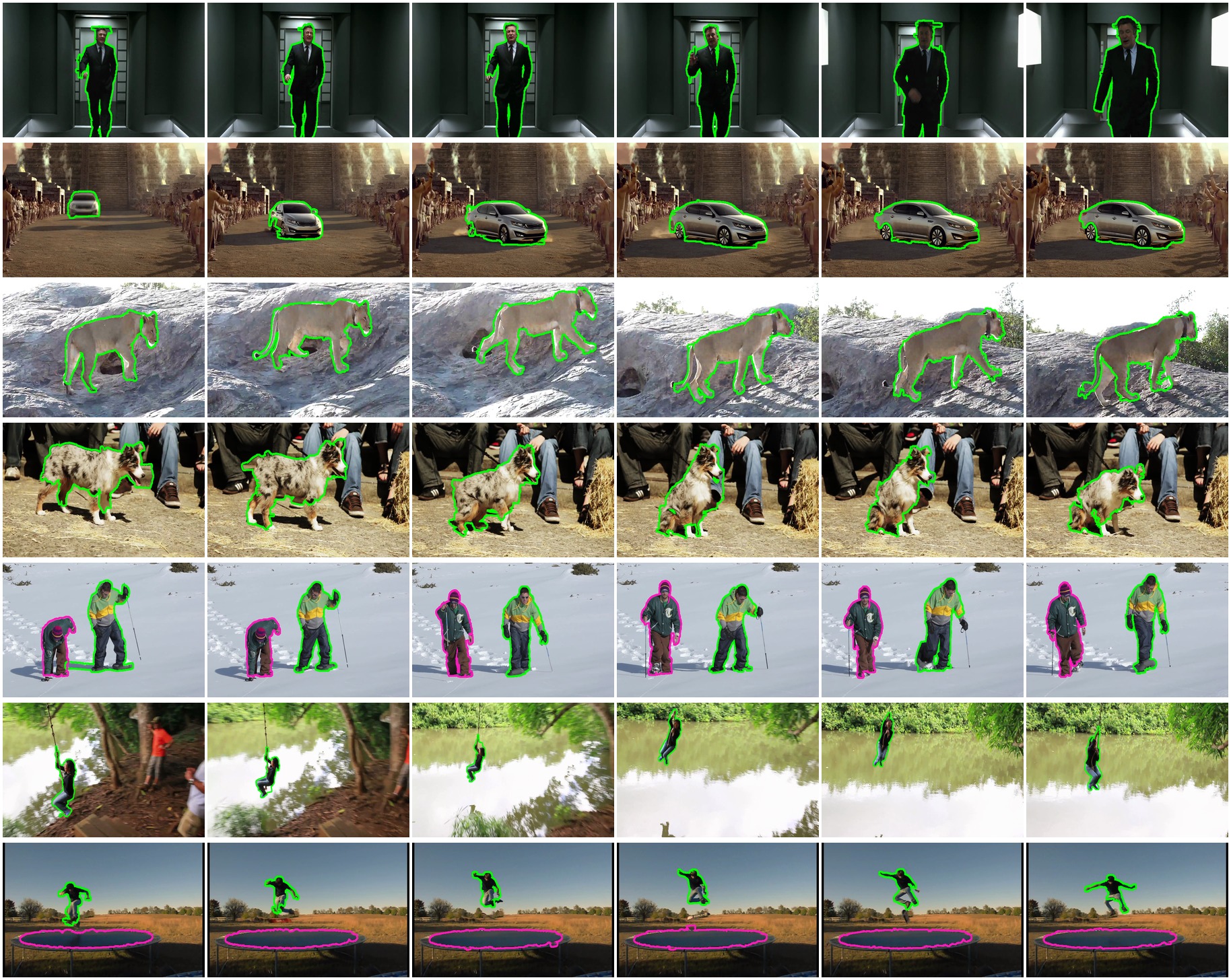}
	\end{tabular}
	\caption{Qualitative results for video segmentation on VSB100 dataset: The results using the propagation method of~\cite{suyog-eccv2014} initialized from the segmentation in the left-most image. This initialization is obtained using our Click Carving method with static and motion-based proposals. Best viewed on pdf. }
	\label{fig:videoseg_result_vsb}
\end{figure*}

\vspace{5pt}
\noindent {\bf Qualitative results on video segmentation propagation:} Figure~\ref{fig:videoseg_result_segtrack} -~\ref{fig:videoseg_result_vsb} shows some qualitative results for video segmentation propagation on the 3 datasets that we used in our experiments. The left-most image in each row shows the best region proposal chosen by a human annotator using Click Carving. Subsequent images show the results of segmentation propagation, when initialized from this selected proposal.

\vspace{5pt}
\noindent  {\bf Conclusion:} We presented a novel interactive video object segmentation technique, \emph{Click Carving} using which only a few clicks are required to obtain accurate spatio-temporal object segmentation in videos. Our method strikes an excellent balance between accuracy and human effort resulting in large savings. Because of the ease of use even for non-experts, our method offers great promise for scaling up video segmentation which can be beneficial for several research communities.

\vspace{5pt}
\noindent  {\bf Acknowledgements:} We would like to thank Naveen Shankar Nagaraja for providing us with the user annotation timing data on the iVideoSeg dataset for comparison. This research is supported in part by ONR YIP N00014-12-1-0754.

\bibliographystyle{IEEEtran}
\bibliography{click_carving}

\end{document}